\documentclass[11pt]{article}
\usepackage{coling2020}
\usepackage{times}
\usepackage{url}
\usepackage{latexsym}

\newcommand{\ACRO}[1]{\textsc{#1}}
\newcommand{\ARRAU}{\ACRO{arrau}}
\newcommand{\BASHI}{\ACRO{bashi}}
\newcommand{\BERT}{\ACRO{bert}}
\newcommand{\CRAC}{\ACRO{crac}}
\newcommand{\FFNN}{\ACRO{ffnn}}
\newcommand{\ISNOTES}{\ACRO{isnotes}}
\newcommand{\RST}{\ACRO{rst}}
\newcommand{\TRAINS}{\ACRO{trains}}
\newcommand{\PEAR}{\ACRO{pear}}
\newcommand{\SCICORP}{\ACRO{scicorp}}
\newcommand{\AR}{anaphor recognition}
\newcommand{\FBR}{full bridging resolution}
\newcommand{\AS}{antecedent selection}

\newenvironment{EXAMPLE}{\begin{list}{}
    {\topsep      4pt
     \itemsep     .0ex
     \labelwidth  25pt
     \leftmargin  30pt
     
     }}{\end{list}}
\newcommand{\ENEW}[1]{\refstepcounter{equation}
\label{#1}\item[(\theequation)\hspace{10pt}]}

\newcounter{EEXAMPLE}

\newenvironment{AEXAMPLE}{\begin{list}{}
    {\topsep      0pt
     \partopsep   0pt
     \itemsep     .0ex
     \labelwidth  2pt
     \leftmargin  0pt
     
     \usecounter{EEXAMPLE}
     }\vskip-\lastskip}{\end{list}}
\newcommand{\ENUMA}[1]{\refstepcounter{equation}
\label{#1}\item[(\theequation)] \begin{AEXAMPLE}}
\newcommand{\ENDENUMA}{\end{AEXAMPLE}}
\newcommand{\SREF}[1]{(\ref{#1})}

\usepackage{amsmath}
\usepackage{booktabs}
\usepackage{footnote}

\usepackage{multirow}
\usepackage{graphicx}
\usepackage{subcaption}

\colingfinalcopy 

\title{Multitask Learning-Based Neural Bridging Reference Resolution}

\author{Juntao Yu \and Massimo Poesio \\
  Queen Mary University of London \\
  {\tt \{juntao.yu, m.poesio\}@qmul.ac.uk} \\}

\date{}

\begin{document}
\maketitle
\begin{abstract}
We propose a multi task learning-based neural model for 
resolving bridging references 
tackling two key challenges.
The first challenge is the lack of large corpora annotated with bridging references.
To address this, we use multi-task learning to help bridging reference resolution with coreference resolution. 
We show that substantial improvements of up to 8 p.p. can be achieved on {\FBR}
with this architecture. 
The second challenge is 
the different definitions of bridging used in
different corpora,
meaning that 
hand-coded systems or systems using special features 
designed for one corpus do not work well with other corpora. 
Our neural model  only uses a small number of corpus independent features, thus can be applied  to different corpora. 
Evaluations with very different bridging corpora ({\ARRAU}, {\ISNOTES}, {\BASHI} and {\SCICORP}) suggest that our architecture
works equally well on all corpora, 
and achieves the SoTA results on {\FBR} for all corpora, 
outperforming the best reported results by up to 36.3 p.p..\footnote{The code is available at 
\url{https://github.com/juntaoy/dali-bridging}
} 
\end{abstract}

\section{Introduction}
\textbf{Anaphora resolution} \cite{karttunen:76,webber:thesis,kamp&reyle:93,garnham:book,poesio-stuckardt-versley:book}
is the aspect of language interpretation concerned with linking nominal expressions to  entities in the context of interpretation (or \textbf{discourse model}). 
As illustrated by
\SREF{ex:anaphora} (adapted from \cite{hou-et-al:CL18}), nominal expressions can be linked to the context in several ways: 
\textbf{corefererence} (linking [The Bakersfield Supermarket], [The business],  [its]),  
\textbf{bridging or associative reference} (linking [the customers] to [the supermarket]) \cite{clark-1975-bridging,prince:81,poesio-etal-2004-learning,hou-et-al:CL18}, and \textbf{discourse deixis} (linking [the murder] to the event of murdering) 
\cite{webber:91,kolhatkar-et-al:CL18}.

\begin{EXAMPLE}
\ENEW{ex:anaphora} 
[The Bakersfield Supermarket] went bankrupt last May.
[The business] closed when [[its] old owner] was murdered by [robbers].
[The murder] saddened [the customers].
\end{EXAMPLE}
\textbf{Bridging reference resolution} is the sub-task of anaphora resolution concerned with identifying and resolving bridging references, i.e., anaphoric references to non-identical associated antecedents. 
Bridging resolution is much less studied 
than the closely related sub-task of coreference resolution, 
which has received a lot of attention (\cite{pradhan2012conllst,wiseman2015learning,lee2017end,lee2018higher}, to mention just a few recent proposals).
One reason
for this is the lack of training data.
Several corpora have been annotated with bridging reference, including e.g.
\ACRO{gnome} \cite{poesio:ACL_discanno04},
{\ISNOTES} \cite{markert-etal-2012-isnotes}, 
{\SCICORP} \cite{roesiger-2016-scicorp} and
{\BASHI} \cite{rosiger-2018-bashi}, 
but they are all rather small, with at most around 1k examples of bridging reference. 
{\ARRAU} \cite{poesio_anaphoric_2008,uryupina-et-al:NLEJ} is much larger, 
but still contains only 5.5k bridging pairs.
It is challenging to train a learning based system on that 
amount of data, particularly the new neural models. 
As a result, the current SoTA systems for {\FBR} are still rule-based, employing a number of heuristic rules many of which are corpus-dependent \cite{hou-etal-2014-emnlp,roesiger-etal-2018-bridging}. 
This is problematic at the light of the 
second challenge for work in this area:
namely, that the definitions of bridging are different in these different 
corpora \cite{roesiger-etal-2018-bridging}. 
Existing corpora differ in whether they attempt to annotate only what Roesiger et al call \textbf{referential bridging} (as in  {\ISNOTES}), or 
the full range of bridging references,
as in {\ARRAU}.%
\footnote{Roesiger et al use `referential bridging' for
the cases in which the bridging reference needs an antecedent in order to be interpretable, such as \underline{the door} in \textit{John walked towards the house. \underline{The door} was open.}.
`Lexical bridging' is when
the bridging reference could also be interpreted autonomously, such as \underline{Madrid} in \textit{I went to Spain last year. I particularly liked \underline{Madrid}.} 
See \cite{poesio:ACL_discanno04,baumann&riester:12,markert-etal-2012-isnotes,hou-et-al:CL18,uryupina-et-al:NLEJ} for a detailed discussion of the annotation schemes, 
and  \cite{roesiger-etal-2018-bridging,roesiger-2018-crac} for a discussion of the implications.} 
The {\ISNOTES}, {\BASHI} and {\SCICORP} corpora consist mostly of referential bridging examples, 
while the {\ARRAU} corpus contains both types of bridging references. 
As a consequence, a system designed for one corpus (e.g. {\ISNOTES}) works poorly when applied to other corpora (e.g. {\ARRAU}), and significant  modifications are needed to make the system works equally well on different corpora \cite{roesiger-2018-crac}.

To tackle these challenges, 
we introduce a multi task learning-based neural model that learns bridging resolution together with coreference resolution.
Multi task architectures have proven effective at exploiting the synergies between distinct but related tasks to in cases when only limited amounts of data are available for one or more of the tasks  \cite{clark-etal-2019-bam}.
Such an architecture should therefore be especially suited for our context, 
given that, linguistically, bridging reference resolution and coreference resolution are two distinct but closely related aspects of anaphora resolution, and indeed were often tackled together in early 
systems \cite{sidner:thesis,vieira&poesio:CL}.
Using a neural network-based approach that minimises  feature engineering enables the system to be more flexible on the choices of corpora. 
We mainly evaluate our system on the {\RST} portion of the {\ARRAU} corpus since it is the largest available resource, 
but we additionally evaluate it 
on the {\TRAINS} and {\PEAR} portion of the {\ARRAU} corpus, {\ISNOTES}, {\BASHI} and {\SCICORP} corpus to demonstrate its tolerance of diversity.

We start with a strong baseline for bridging adapted from the SoTA coreference architecture \cite{lee2018higher,kantor-globerson-2019-coreference} 
enhanced by 
{\BERT} embeddings \cite{devlin2019bert}. 
We extend the system to  multi-task learning by adding a coreference classifier that shares part of the network with the bridging classifier. 
In this way, we improve
{\FBR} and its subtasks ({\AR} and {\AS})  by 6.5-7.3 p.p. respectively. 
But because 
the number of coreference examples is much larger 
than the number of
bridging pairs, the dataset is highly imbalanced. 
We achieve
further improvements of 1.7 p.p. and 6.6 p.p. on {\FBR} and {\AR} 
by using undersampling during the training. 
This final system achieves  SoTA results on both {\FBR} and its subtasks, i.e. 4.5, 6 and 9.5 p.p. higher than the best reported results \cite{roesiger-2018-crac} on {\FBR}, {\AR} and {\AS} respectively.  
Evaluation on {\TRAINS}, {\PEAR}, 
{\ISNOTES}, {\BASHI} and {\SCICORP}
shows the same trend. 
Although 
the datasets 
are much smaller and the annotation schemes for {\ISNOTES}, {\BASHI} and {\SCICORP} are different from {\ARRAU}, our system works equally well, 
achieving
the new SoTA results on {\FBR} and {\AR} for all six corpora as well as five corpora on {\AS}.

\section{Related Work}

\subsection{Bridging Reference Resolution}
Bridging reference resolution involves two subtasks: {\AR} and {\AS} \cite{hou-et-al:CL18}. 
Early work on bridging resolution mostly focused on 
definite bridging anaphors \cite{sidner:thesis,vieira&poesio:CL,lassalle&denis:DAARC11},
but later systems covered 
unrestricted {\AS} 
\cite{poesio-etal-2004-learning,hou-etal-2013-global}.
\newcite{hou-etal-2013-global} 
introduced a model based on
Markov logic networks 
and using an extensive set of features and constraints.
They evaluated the system with both local and global features on  {\ISNOTES}, and showed that  global features can greatly improve performance. 
The system was later extended in \cite{hou-2018-naacl,hou-2018-emnlp,hou-et-al:CL18} 
to explore additional features from  embeddings tailored for bridging resolution, to advanced antecedent candidate selection using 
the Penn Discourse Treebank (\textit{d-scope-salience}). 
Recently, \newcite{hou-2020-acl} framed the {\AS} task as question answering, and pre-trains the system with a large synthetic bridging corpus; this system achieves  SoTA results on {\ISNOTES}. 
But those systems are highly specialized for
the {\ISNOTES} corpus, hence perform less well on other corpora.
The {\AR} subtask is usually solved as a part of the information status task \cite{markert-etal-2012-isnotes,hou-2016-incremental,hou-et-al:CL18}.

Recent systems for {\FBR} 
include 
\cite{hou-etal-2014-emnlp,hou-et-al:CL18,roesiger-etal-2018-bridging,roesiger-2018-crac}. 
\newcite{hou-etal-2014-emnlp} proposed a rule-based system for {\FBR} with the {\ISNOTES} corpus, consisting of a rich system of rules motivated by linguistic knowledge. 
They also evaluated a learning-based system that uses the rules as features, but the learning-based system only outperforms the rule-based system's F1 score by 0.1 percentage points. 
The rule-based system was
later adapted by \newcite{roesiger-etal-2018-bridging}; \newcite{roesiger-2018-crac} 
for
{\FBR} on  {\ARRAU},
but since  {\ARRAU}  follows a more general 
definition of bridging, 
most of the rules had to be changed. 

\subsection{Multi-task Learning for Under-Resourced Tasks}
Multi-task learning 
has been successfully used in several NLP applications
\cite{collobert2008unified,luong2015multi,kiperwasser2018scheduled,clark-etal-2019-bam}.  
Normally, the goal of multi-task learning is  to improve performance on all  tasks;
but 
in an under-resourced setting, the aim often is only to improve performance on the low resource task/language/domains (the \textbf{target task}). 
This is sometimes known as  \textbf{shared representation based transfer learning}. 
\newcite{yang2017transfer} applied transfer learning to sequence labelling tasks; the deep hierarchical
recurrent neural network used in their work is fully/partially shared between the source and the target tasks. They demonstrated that SoTA performance can be achieved by using models trained on multi-tasks. 
\newcite{cotterell2017low} trained a neural NER system on 
a combination of high-/low-resource languages to improve  NER for the low-resource languages. 
In their work,  character-based embeddings are shared across the languages. 
Recently, \newcite{zhou2019dual} introduced a multi-task network together with adversarial learning for under-resourced NER. The evaluation on both cross-language and cross-domain settings shows that partially sharing the BiLSTM works better for cross-language transfer, while for cross-domain setting, the system performs better when the LSTM layers are fully shared.

\subsection{Neural Coreference Resolution}
By contrast with bridging reference, 
coreference resolution
has been extensively studied. 
\newcite{wiseman2015learning,wiseman2016learning} first introduced a neural network-based approach to solving coreference in a non-linear way. 
\newcite{clark2016improving} integrated reinforcement learning to let the model, optimized directly on the B$^3$ scores. 
\newcite{lee2017end}  proposed a neural joint approach for mention detection and coreference resolution. Their model does not rely on parse trees; instead, the system learns to detect mentions by exploring the outputs of a BiLSTM. 
After the introduction of  context dependent word embeddings such as ELMo \cite{peters2018elmo} and {\BERT} \cite{devlin2019bert}, the \newcite{lee2017end} system has been greatly improved by those embeddings \cite{lee2018higher,kantor-globerson-2019-coreference} to achieve  SoTA results. 
We use a simplified version of the model by \cite{lee2018higher,kantor-globerson-2019-coreference}  
as baseline. 

\section{Methods}
\subsection{The Single-Task Baseline System}

We use  as our single-task baseline for bridging reference a simplified version of the SoTA coreference systems by  \newcite{lee2018higher,kantor-globerson-2019-coreference}, 
since  bridging resolution  is closely related to  coreference: like coreference it requires establishing a link to an entity in the discourse model, but through a non-identity relation. 
The \newcite{kantor-globerson-2019-coreference} model is an extended version of  \cite{lee2018higher}; the main difference is that 
Kantor et al use
{\BERT} embeddings \cite{devlin2019bert} instead of the ELMo embeddings \cite{peters2018elmo} used by
Lee et al. 
These systems have similar architecture and both do  mention detection and coreference jointly. 
We only use the coreference part of the system, since for bridging resolution evaluation is usually  on 
gold mentions. 

Our baseline system first creates representations for mentions using the output of a BiLSTM. 
The sentences of a document are encoded from both directions to obtain a representation for each token in the sentence. 
The BiLSTM takes as input the concatenated embeddings ($(emb_t)_{t=1}^{T}$) of both word and character levels.
For word embeddings,  GloVe \cite{pennington2014glove} and {\BERT} \cite{devlin2019bert} embeddings are used. 
Character embeddings are learned from a convolution neural network (CNN) during training. 
The tokens are represented by concatenated outputs from the forward and the backward LSTMs. 
The token representations $(x_t)_{t=1}^{T}$  are used together with head representations ($h_i$) to represent mentions ($M_i$). 
The $h_i$ of a mention is obtained by applying 
attention over its token representations ($\{x_{b_i}, ..., x_{e_i}\}$), where $b_i$ and $e_i$ are the indices of the start and the end of the mention respectively. 
Formally, we compute $h_i$, $M_i$ as follows:

\vspace{-10pt}
$$\alpha_t = \textsc{ffnn}_{\alpha}(x_{t})\ ,\ a_{i,t} = \frac{exp(\alpha_t)}{\sum^{e_i}_{k=b_i} exp(\alpha_k)}$$
$$ h_i = \sum^{e_i}_{t=b_i} a_{i,t} \cdot x_t\ ,\ M_i = [x_{b_i}, x_{e_i},h_i,\phi(i)]$$
where $\phi(i)$ is the mention width feature embeddings. 
Next, we pair the mentions with candidate antecedents to create a pair representation 
($P_{(i,j)}$):

\vspace{-8pt}
$$P_{(i,j)} = [M_i, M_j, M_i \circ M_j,\phi(i,j)]$$
where $M_i$, $M_j$ is the representation of the antecedent and anaphor respectively, 
$\circ$ denotes  element-wise product, and
$\phi(i,j)$ is the distance feature between a mention pair. 
To make the model computationally tractable, we consider for each mention a maximum 150 candidate antecedents\footnote{The number of  maximum antecedents was tuned on the dev set.}.

The next step is to compute the pairwise score ($s(i,j)$). Following \newcite{lee2018higher}, we add an artificial antecedent $\epsilon$ to deal with cases of non-bridging anaphor mentions or cases when the antecedent does not appear in the candidate list. We compute $s(i,j)$ as follows:

\vspace{-8pt}
$$
s(i,j) = \Bigg\{
  \begin{tabular}{ll}
  0& $i=\epsilon$ \\
  $\textsc{ffnn}(P_{(i,j)})$& $i\neq \epsilon$
  \end{tabular}
$$

For each mention the predicted antecedent is the one has the highest $s(i,j)$, a bridging link will be only created if the predicted antecedent is not $\epsilon$. 

\subsection{Our Multi-task Learning Architecture}
Choosing a source task that is closely related to bridging resolution is crucial to the success of our multi-task learning model.
In this work, we use coreference as the source task.
The key intuitions behind this choice 
are: 
(i)  from a language interpretation point of view, resolving anaphoric coreference and anaphoric bridging reference are closely related tasks in that they both
involve trying to identify relations between anaphors and antecedents \cite{poesio:ana_book_linguistic}--indeed, the two tasks were typically tackled jointly by non ML-based anaphora resolution systems \cite{sidner:thesis,hobbs-et-al:abduction,vieira&poesio:CL}; 
(ii) from the point of view of the model, both  tasks rely on a good mention representations and can be solved by neural mention pair models.

\begin{figure}[t]
    \centering
    \includegraphics[width=.95\columnwidth]{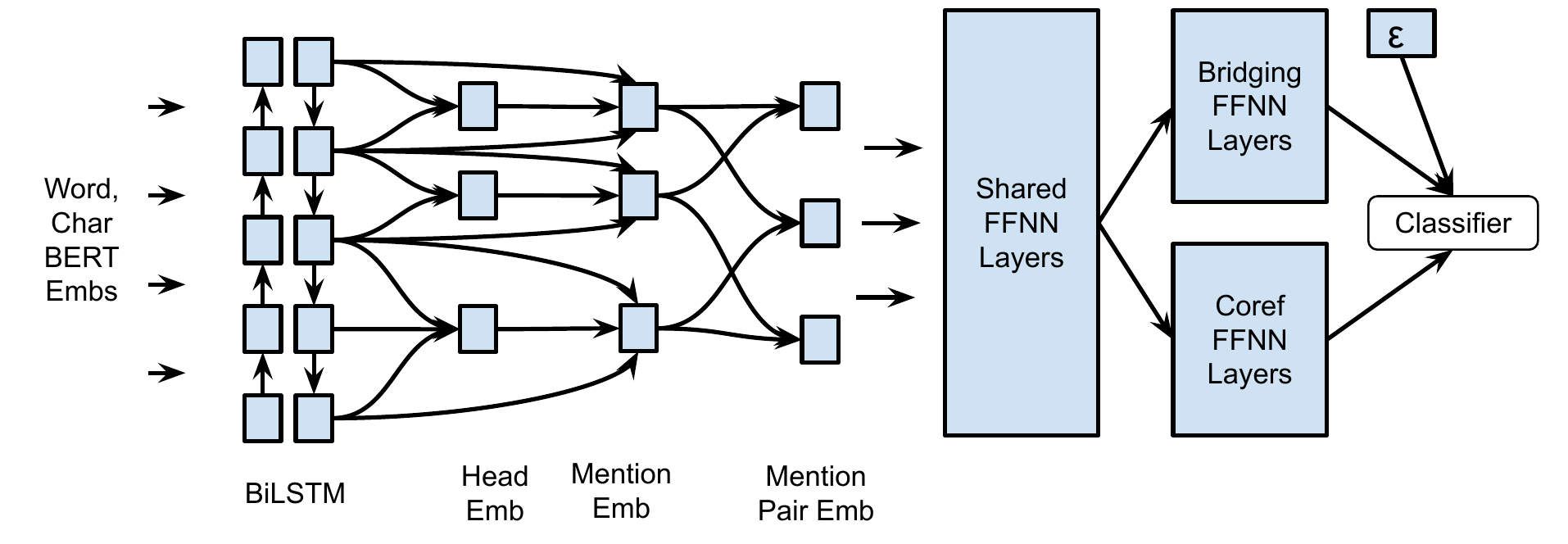}
    \caption{The proposed multi-task architecture.
    }
    \label{fig:nngraph}
\end{figure}

We turn our model 
into a multi-task model 
by adding to the architecture an additional classifier for coreference, and jointly predicting coreference and bridging (Figure \ref{fig:nngraph}).
We use the same candidate antecedents for both bridging and coreference tasks. 
As shown in Figure \ref{fig:nngraph}, our model uses shared mention representations (i.e. the word embeddings and the BiLSTM) with additional options to share some/all hidden layers of the \textsc{ffnn}. 
By sharing most of the network structure, the mention representations learned by the coreference task become accessible by bridging resolution.

\subsection{Learning with Imbalanced Data}
Following \newcite{lee2018higher} we optimise our system on the marginal log-likelihood of all correct antecedents. For bridging, we consider a bridging antecedent correct if it is from the same gold coreference cluster \textsc{gold}$(i)$ of the gold bridging antecedent. For coreference, the correct antecedents is implied from the gold coreference cluster \textsc{gold}$(j)$ the mention belongs to. We compute both bridging and coreference losses 
as follows:

\vspace{-8pt}
$$log \prod_{j=1}^{N}\sum_{\hat{y} \in Y(j) \cap \textsc{gold}(i/j)} s(\hat{y},j)$$
in case mention $i$ is not a bridging/coreference anaphor or $Y(j)$ (the candidate antecedents) does not contain mentions from $\textsc{gold}(i)$ for bridging or \textsc{gold}$(j)$ for coreference, we set \textsc{gold}$(i/j) = \{\epsilon\}$.

When training with coreference one of the problem we need to face is  class imbalance. 
Consider the {\RST}  portion of {\ARRAU} as an example
(mostly WSJ text). 
The corpus contains 72k mentions in total: of these, 
45k (63\%) are discourse-new  (DN), 
24k (33\%)  are discourse-old  (DO),
and the remaining 3k  (4\%)  are bridging anaphors. 
Training the model on such an imbalanced corpus may significantly harm  recall with bridging anaphors. 

To reduce the negative effect of this imbalance, we use undersampling during to training by randomly removing DN and DO examples to make the corpus more balanced. 
More precisely, we use a heuristic negative example ratio $\gamma$ to control the total number of negative examples during the training, so that, e.g.,  $\gamma = 2$ means we keep 6k DN and 6k DO examples during training. 
We set a value for $\gamma$ 
by trying a few small values in  preliminary experiments;
they all gave very similar results, hence we set  $\gamma = 2$ in the  experiments below.

\section{Experiments}
\begin{table}[t]
\centering
\resizebox{0.95\textwidth}!{
\begin{tabular}{lllll}
\toprule

Corpus&Genre&Bridging Type&Mention Type&Number of Bridging\\\midrule
{\ARRAU} {\RST}&WSJ news&lexical, referential bridging&Gold&3303\\ 
{\ARRAU} {\TRAINS}&Dialogues&lexical, referential bridging&Gold&558\\
{\ARRAU} {\PEAR}&Narrative&lexical, referential bridging&Gold&303\\
{\ISNOTES}&WSJ news&referential bridging&Gold&663\\
{\BASHI}&WSJ news&referential bridging&Predicted&459\\
{\SCICORP}&Scientific texts&referential bridging&Predicted&1366\\
\bottomrule
\end{tabular}}
\caption{\label{table:corpus_bridging} The statistics of bridging corpora used in our experiments. }
\end{table}

\textbf{Datasets} We evaluated our systems on {\ARRAU} \cite{poesio_anaphoric_2008,uryupina-et-al:NLEJ}, {\ISNOTES} \cite{markert-etal-2012-isnotes}, {\BASHI} \cite{rosiger-2018-bashi} and {\SCICORP} \cite{roesiger-2016-scicorp} with  {\ARRAU} {\RST} as our primary dataset as it is substantially larger than other datasets (see Table \ref{table:corpus_bridging} for more detail).

Bridging references are annotated in {\ARRAU} according to the scheme in \cite{uryupina-et-al:NLEJ}, which
covers both what \newcite{roesiger-2018-crac} call `lexical' and  `referential' bridging. 
The corpus was used for 
Task 2 of 
the {\CRAC} 2018 shared task \cite{poesio2018crac},
focused on bridging resolution. 
As done in the {\CRAC} shared task, we evaluate our system on all three subcorpus {\RST},{\TRAINS} and {\PEAR} stories\footnote{We followed \newcite{roesiger-2018-crac} in excluding a small portion of bridging links that are problematic (e.g. empty antecedent, split antecedent).}. The {\RST} portion of the corpus, consisting of 413 news documents (1/3 of the WSJ section of the Penn Treebank). 
We used the default train/dev and test subdivisions. The {\TRAINS} and {\PEAR} portion of the corpus contains 114 dialogues and 20 fictions respectively. Since the {\TRAINS} and {\PEAR} are much smaller we use 10-fold cross validation and report the results on test set to compare with previous work.

The {\ISNOTES} corpus consists of 50 documents  from the WSJ portion of \ACRO{OntoNotes}, with 663 bridging pairs annotated as well as  fine-grained information status according to the scheme in \cite{markert-etal-2012-isnotes}. 
Bridging is annotated as one of the information status subclasses. 
Like {\ISNOTES}, the {\BASHI} corpus contains 50 documents from the WSJ portion of \ACRO{OntoNotes}. The dataset has 459 bridging pairs annotated according to a annotation scheme similar to that of the {\ISNOTES} corpus \cite{rosiger-2018-bashi}. We follow \newcite{hou-2020-acl} to mix use the {\ISNOTES} and {\BASHI} corpora\footnote{For each fold of the cross-validation, we train the system with 90\% of the main corpus plus the 
full auxiliary corpus and tested on the rest 10\% of the main corpus.}.
The {\SCICORP} corpus uses text from a very different  domain of scientific texts. 
The corpus has in total 1366 bridging pairs annotated, again according to its own annotation scheme \cite{roesiger-2016-scicorp}. 
Since those corpora are rather small, we used 10-fold cross-validation to evaluate 
on them. 

\textbf{Evaluation metrics} Following, 
e.g.,  Hou et al., we evaluate 
on both {\FBR} and its subtasks ({\AR}/{\AS}). For {\FBR} and {\AR} we report F1 scores.\footnote{Following \cite{roesiger-2018-crac} we consider a predicted bridging antecedent is correct when it belongs to the same gold corefernce cluster as the gold bridging antecedent.} 
For {\AS} we report accuracy as it uses gold bridging anaphors.

\textbf{Hyperparameters} Apart from the two parameters introduced by our model (maximum number of antecedents and negative example ratio $\lambda$), we use  the default settings from \newcite{lee2018higher}, and replace their ELMo settings with the {\BERT} settings from \newcite{kantor-globerson-2019-coreference}. 
 Table \ref{table:config} summarizesthe hyperparameter settings of our model. 
We train the models evaluated on the {\ARRAU} {\RST} for 200 epochs, and for 50 epochs the models trained on the other corpora.

\begin{table}[t]
    \centering
    \resizebox{0.5\textwidth}!{
    \begin{tabular}{l l}
    \toprule
    \bf Parameter & \bf Value \\
    \midrule
    BiLSTM layers, size, dropout&3, 200, 0.4\\
    FFNN layers, size, dropout& 2, 150, 0.2\\
    CNN filter widths, size& [3,4,5], 50\\
    Char,GloVe,Feature embedding size&8, 300, 20\\
    BERT layer, size& Last 4, 1024 \\
    Embedding dropout & 0.5\\
    Max num of antecedent&150\\
    Negative example ratio ($\gamma$) &2\\
    Optimiser, Learning rate & Adam, 1e-3\\
    \bottomrule
    \end{tabular}}
    \caption{Hyperparameters for our models.}
    \label{table:config}
\end{table}

\section{Results and Discussions}
\begin{table}[t]
\centering
\begin{subtable}{\linewidth}
\centering
\resizebox{0.5\textwidth}!{
\begin{tabular}{lrll}
\toprule
System&Shared Network&{\RST}&{\ISNOTES}\\\midrule
bridging only&&47.4&33.8\\ \midrule
&embeddings, LSTM&50.9&38.7\\
multi-task&+ 1 {\FFNN} Layer&\bf 54.7&\bf 43.7\\ 
&+ 2 {\FFNN} Layer& 51.7&40.1\\
\bottomrule
\end{tabular}}
\caption{\label{table:dev_ante_selection} {\AS}}
\end{subtable}
\begin{subtable}{\linewidth}
\centering
\resizebox{0.95\textwidth}!{
\begin{tabular}{lcccccccccccc}
\toprule
&\multicolumn{6}{c}{\RST}&\multicolumn{6}{c}{\ISNOTES}\\\cmidrule{2-13}
&\multicolumn{3}{c}{anaphor rec.}&\multicolumn{3}{c}{full bridging res.}&\multicolumn{3}{c}{anaphor rec.}&\multicolumn{3}{c}{full bridging res.}\\\cmidrule{2-13}
System&P&R&F1&P&R&F1&P&R&F1&P&R&F1\\\midrule
bridging only&50.0&12.5&20.0&34.5&8.6&13.8& 63.9&16.2&25.8&33.3&8.5&13.5\\ \midrule
multi-task&47.3&19.0&27.1&35.5&14.2&20.3& 59.3&22.5&32.7&31.5&12.0&17.4\\
\ + undersampling&31.5&36.2&\bf 33.7&20.6&23.7&\bf 22.0& 45.6&47.2&\bf 46.4&19.1&19.7&\bf 19.4\\
\bottomrule
\end{tabular}
}
\caption{\label{table:dev_full_bridging} {\FBR}}
\end{subtable}
\caption{Parameter tuning
on the dev set of {\ARRAU} {\RST} and {\ISNOTES}.}
\end{table}

\subsection{Evaluation on the {\ARRAU} {\RST} corpus}
We first evaluated our multi-task learning based system on the {\AS} subtask, to assess the suitability of our model for bridging. 
The {\AS} subtask uses 
gold bridging anaphors, hence it is simpler 
than 
{\FBR} which additionally involves identifying the bridging references. 
Focusing on a simpler task also allowed us to have a clearer view of the effects of multi-task learning. 
In this experiment, we configured the system to share only the mention representations (the word embeddings and BiLSTM). 
As illustrated in Table \ref{table:dev_ante_selection}, the baseline system already achieved a pretty good accuracy for this type of task. 
Although starting from a strong baseline, our multi-task learning based system achieved an 
improvement of 3.5 percentage points, confirming our hypothesis that  coreference is a good source task for bridging.

\textbf{Sharing The Feed-forward Layers} We further extended our model 
to share the {\FFNN} layers in addition to the mention representations. 
The {\FFNN} layers have access to pairwise representations to learn the relations between the anaphors and the antecedents, hence 
contain
useful information regarding how likely two mentions are to be related. 
As expected, this additional sharing of the {\FFNN} layers 
resulted in additional
improvements. 
A further improvement of 3.8 percentage points was achieved by sharing  1 additional {\FFNN} layer. 
The accuracy drops when
both   hidden layers are shared between coreference and bridging, but the performance is still higher when compared with the model that only shares  mention representations. Overall, the multi-task model achieved a substantial gain of 7.3 percentage points when compared with the 
system only carrying out bridging reference resolution
(see Table \ref{table:dev_ante_selection}).

\begin{table}[t]
\centering
\resizebox{.5\textwidth}!{ 
\begin{tabular}{llll}
\toprule
&Number&Baseline&Multi-task\\\midrule
\ACRO{subset} &113&49.6&57.5\\
\ACRO{element}&89&50.6&52.8\\
\ACRO{poss}   &8&0.0&12.5\\
\ACRO{other}  &11&54.6&54.6\\
\ACRO{undersp-rel}&11&27.3&72.7\\
\bottomrule
\end{tabular}}
\caption{\label{table:arrau_dev_ante_selection_analysis} The accuracy comparison between our best multi-task model and the baseline on {\AS} for different bridging relations. Evaluated on the dev set of {\ARRAU} {\RST}}
\end{table}

The {\ARRAU} {\RST} corpus also contains information about the semantic relation between the bridging reference and its antecedent. Five different relations are annotated: subset (\ACRO{subset}), set membership (\ACRO{element}), a generalised possession relation including part-of relations (\ACRO{poss}), \textit{other} NPs such as \underline{the other} in \textit{Two men entered the pub. One man ordered a beer, \underline{the other} a glass of wine} (\ACRO{other}) and bridging relations that do not fit into any defined classes (\ACRO{undersp-rel}). We further compare our best multi-task model with the single-task baseline on different bridging relations to find out which relation was helped the most by  multi-task learning. Table \ref{table:arrau_dev_ante_selection_analysis} shows the result of our comparison on the development set. 
The development set contains mostly bridging references of type \ACRO{element} and \ACRO{subset}, and  a small number of the other three relations. For the two main classes, the multi-task system improved strongly the \ACRO{subset} bridges with a large gain of 8\%. The improvement on \ACRO{element} bridges is smaller (2.2\%). One possible explanation for the larger improvement with \ACRO{subset} is that many \ACRO{subset} bridging references in {\ARRAU} {\RST} denote a subtype of a previously introduced type, as in \textit{computers ... \underline{personal computers}}, and knowledge about coreference may be especially helpful for such cases.

\textbf{Full Bridging Resolution} 
Having ascertained the benefits of our multi-task model for {\AS}, we applied the best settings (sharing mention representations and the 1st hidden layer of the {\FFNN}) to {\FBR} as well. 
We also report the F1 scores for bridging {\AR}, a byproduct of {\FBR}. 
Table \ref{table:dev_full_bridging} shows a comparison between the single-task baseline and the multi-task models. 
The baseline model trained without multi-task learning achieved F1 scores of 13.8\% and 20\% on {\FBR} and {\AR}, respectively. 
The low F1 scores are mainly due to a poor recall in both tasks, a well known problem with bridging reference resolution. 
When applying  multi-task learning, the F1 scores improve substantially (6.5\% and 7.1\% for {\FBR} and {\AR} respectively). 
These F1 improvements  are mainly  a result of better recall; 
the precisions of the two models are similar. 
This suggests that learning with coreference does help the model to capture more correct bridging pairs. 
However, recall is still much lower than precision.
As this might a result of  data imbalance, 
we  applied  undersampling during training, to train the model on a more balanced dataset. 
As shown in Table \ref{table:dev_full_bridging}, with undersampling the model has a more balanced precision and recall, and also achieves better F1 scores on both {\FBR} and {\AR}. 
The new model achieved improvements of 6.6\% and 1.7\% on {\AR} and {\FBR}, respectively, when compared with the model without undersampling.
Overall, our multi-task models showed their merit on both tasks and achieved considerable gains of 8.2\% and 13.7\% when compared with the single-task system.

\begin{savenotes}
\begin{table}[t]
\centering
\resizebox{0.7\textwidth}!{
\begin{tabular}{lcccccc}
\toprule
&{\RST}&{\TRAINS}&{\PEAR}&{\ISNOTES}\footnote{\newcite{hou-et-al:CL18} report a score of 50.7\%, but their system relies on hand-coded information from the Penn Discourse Treebank for antecedent candidate selection, so that systems is not really comparable to systems using only automatic information. 
Hence, 
we follow \newcite{hou-2020-acl} and exclude that result from the table.}&{\BASHI}&{\SCICORP}\\\midrule
\newcite{hou-etal-2013-global}&-&-&-&41.3&-&-\\
\newcite{hou-2018-emnlp} &32.4&-&-&46.5&27.4&-\\ 
\newcite{roesiger-2018-crac}&39.8&48.9&28.2&-&-\\
\newcite{hou-2020-acl}&34.6&-&-&\bf 50.1&-&-\\
 \midrule
Our model&\bf 49.3&\bf 50.9&\bf 61.2&43.7&\bf 36.0& \bf 33.4
\\ 
\bottomrule
\end{tabular}}
\caption{\label{table:all_ante_selection} 
Comparing our model with the SoTA for  {\AS}. 
}
\end{table}
\end{savenotes}

\begin{savenotes}
\begin{table}[t]
\centering
\resizebox{.95\textwidth}!{ 
\begin{tabular}{lclcccccc}
\toprule
\multirow{2}{*}{Corpus}&Gold Coreference&\multirow{2}{*}{Models}&\multicolumn{3}{c}{anaphor rec.}&\multicolumn{3}{c}{full bridging res.}\\\cmidrule{4-9}
&Anaphors Setting&&P&R&F1&P&R&F1\\\midrule

\multirow{3}{*}{\RST}&Keep
&Our model&31.8&29.8&30.8&20.2&18.9&19.5\\ \cmidrule{2-9}

&\multirow{2}{*}{Remove}&
\newcite{roesiger-2018-crac}&29.2&32.5&30.7&18.5&20.6&19.5\\
&&Our model&37.6&35.9&\bf 36.7&24.6&23.5&\bf 24.0\\\midrule

\multirow{3}{*}{\TRAINS}&Keep&
Our model&49.4&36.0&41.6&33.7&24.6&28.4\\\cmidrule{2-9}
&\multirow{2}{*}{Remove}&
\newcite{roesiger-2018-crac}&39.3&21.8&24.2&27.1&21.8&24.2\\
&&Our model&62.2&40.4&\bf 48.9&39.2&25.4&\bf 30.9
\\\midrule

\multirow{3}{*}{\PEAR}&Keep&
Our model&74.7&48.8&59.0&68.4&44.6&54.0\\\cmidrule{2-9}
&\multirow{2}{*}{Remove}&
\newcite{roesiger-2018-crac}&75.0&16.0&26.4&57.1&12.2&20.1\\
&&Our model&81.1&49.6&\bf 61.5&74.3&45.5&\bf 56.4\\\midrule

\multirow{6}{*}{\ISNOTES}&
\multirow{3}{*}{Keep}&
\newcite{hou-etal-2014-emnlp}
\footnote{The results of \newcite{hou-etal-2014-emnlp} are 
from \newcite{roesiger-etal-2018-bridging}, as 
they were obtained 
on an unknown subset of the corpus.}&65.9& 14.1& 23.2& 57.7& 10.1& 17.2\\
&&\newcite{roesiger-etal-2018-bridging}&45.9&18.3&26.2&32.0&12.8&18.3\\
&&Our model&53.9&33.6&\bf 41.4&31.6&19.8&\bf 24.3\\ \cmidrule{2-9}

&\multirow{3}{*}{Remove}&
\newcite{roesiger-etal-2018-bridging}&71.6&18.3&29.2&50.0&12.8&20.4\\
&&Our model&58.3&35.1&\bf 43.8&33.5&20.2&\bf 25.2\\\midrule

\multirow{3}{*}{\BASHI}&Keep&
Our model&34.4&34.2&34.3&17.7&17.5&17.6\\ \cmidrule{2-9}
&\multirow{2}{*}{Remove}&
\newcite{roesiger-etal-2018-bridging}&49.4&20.2&28.7&24.3&10.0&14.1\\
&&Our model&35.3&34.9&\bf 35.1&18.2&18.0&\bf 18.1\\\midrule

\multirow{3}{*}{\SCICORP}&Keep&
Our model&45.0&35.7&39.8&21.5&17.1&19.0
\\ \cmidrule{2-9}
&\multirow{2}{*}{Remove}&
\newcite{roesiger-etal-2018-bridging}&17.7&0.9&8.1&3.2&0.9&1.5\\
&&Our model&52.9&41.2&\bf 46.3&25.0&19.4&\bf 21.9
\\
\bottomrule
\end{tabular}}
\caption{\label{table:all_full_bridging} 
Comparing our model with the SoTA for  {\FBR}.%
}
\end{table}
\end{savenotes}

\textbf{Comparison with the State of the Art} 
We then evaluated our model on the test set of {\ARRAU} {\RST} to compare it with the previously reported state of the art on the same dataset. 
Table \ref{table:all_ante_selection} shows the comparison on {\AS}. 
The best reported system on this task, \cite{roesiger-2018-crac},  is a modified version of the original rule-based system designed for {\ISNOTES} by \newcite{hou-etal-2014-emnlp}. 
Our system outperforms the current state of the art by nearly 10 percentage points.  
Table \ref{table:all_full_bridging} presents the comparison on the {\FBR} and {\AR}. 
Since the only reported {\FBR} results on {\ARRAU} \cite{roesiger-2018-crac} are evaluated with gold coreferent anaphors removed, we follow the same method to remove gold coreferent anaphors from the evaluation, but  
we also report the results with coreferent anaphors included for future reference. 
Filtering out the gold coreferent anaphors the task is easier, resulting in better F1 scores. 
After filtering out gold  coreferent anaphors, our system achieved F1 scores of 24\% and 36.7\% on {\FBR} and {\AR} respectively, which is 4.5\% and 6\% higher than the scores reported in \newcite{roesiger-2018-crac}. 
Overall, our model achieved the new SoTA results on both {\FBR} and its subtasks.

\subsection{Evaluation on the {\ARRAU} {\TRAINS} and {\PEAR} corpora}

We then evaluate our system on the {\TRAINS} and {\PEAR} sub-corpora of  {\ARRAU}. 
For both corpora, the only reported results are by \newcite{roesiger-2018-crac}. For {\AS} our system achieved scores 2\% and 33\% better than theirs on {\TRAINS} and {\PEAR} respectively (see Table \ref{table:all_ante_selection}). For the other two tasks, they only report scores after filtering out the gold coreferent anaphors, when evaluate in the same setting, our system achieved substantial improvements of up to 35.1\% and 36.3\% for {\AR} and {\FBR} respectively. Overall, our system is substantially better than the \newcite{roesiger-2018-crac} system on both {\TRAINS} and {\PEAR} corpora.

\subsection{Evaluation on the {\ISNOTES} corpus}
\label{section:evaluation_isnotes}
Most recent work on bridging reference resolution was evaluated on  {\ISNOTES}; 
a number of systems were developed for both {\FBR} \cite{hou-etal-2014-emnlp,roesiger-etal-2018-bridging} and {\AS} \cite{hou-etal-2013-global,hou-2018-naacl,hou-2018-emnlp,hou-2020-acl}. Since the {\ISNOTES} follows a very different annotation scheme than that of  {\ARRAU}, to confirm the suitability of our best setting for {\ARRAU} on corpora only containing referential bridging examples ({\ISNOTES}, {\BASHI} and {\SCICORP}) we run additional parameter tuning on the {\ISNOTES} corpus. For parameter tuning we use the same 10 documents  used by \newcite{roesiger-etal-2018-bridging} as a development set, and use the rest 40 documents for training. As shown in Table \ref{table:dev_ante_selection} and Table \ref{table:dev_full_bridging} the results on {\ISNOTES} follows the same trend as for {\ARRAU} {\RST}, the best settings for two corpora remain the same.

To compare with the SoTA systems, we use  10-fold cross-validation to obtain predictions for the whole corpus. 
On the {\FBR} task, our system outperforms all the previous results  both when coreferent anaphors are included (6\%) and when they are
excluded (4.8\%). 
The improvements on {\AR} are larger, and our system is more than 14\% better in both settings. 

For {\AS}, however, our system achieved a  result 2.4\% better than the best reported system with out the access to large synthetic bridging corpora \cite{hou-etal-2013-global}, but lower than those obtained with the help of such corpora \cite{hou-2018-emnlp,hou-2020-acl}. The synthetic bridging corpora extracted by \newcite{hou-2018-emnlp} and \newcite{hou-2020-acl} are based on the prepositional (e.g., X preposition Y) or the possessive structures (e.g., Y
’s X) which is common in the {\ISNOTES} corpus but not in the other corpus (e.g. {\ARRAU}). Models enhanced by such corpora are overfitted to the {\ISNOTES} corpus, the performances are much lower when evaluated on other corpora. For example, those systems achieved a much lower results on {\ARRAU} {\RST} (14.7\% - 16.9\%) when compared with ours. In additional, both systems rely on the gold semantic information for selecting candidate antecedents during test time, which is less realistic.

Overall, on the {\ISNOTES} corpus our system achieved a competitive result on {\AS} and the SoTA on {\FBR} and {\AR}.


\subsection{Evaluation on the {\BASHI} corpus}
We next compare our system with previous models on the {\BASHI} corpus. Since gold mentions are not annotated in {\BASHI}, we use NPs as our predicted mentions without filtering\footnote{The NPs do not belong to coreference clusters or bridging relations are treated as non-mention during training.}. 
For {\AS} the only reported result on {\BASHI} corpus is from \newcite{hou-2018-emnlp}\footnote{\newcite{hou-2020-acl} only reported results on a subset of the {\BASHI} corpus but not the full corpus.} (see Table \ref{table:all_ante_selection}). Our system achieved a accuracy of 36\%, which is 8.6\% better than that of \newcite{hou-2018-emnlp}. 
 \newcite{roesiger-etal-2018-bridging} reported the only results for  {\FBR} and {\AR}. 
 Our system achieves F1 scores that are 6.4\% and 4\% better than their results on {\AR} and {\FBR} respectively (see Table \ref{table:all_full_bridging}). 
 Overall, our model achieves new SoTA on all three tasks.


\subsection{Evaluation on the {\SCICORP} corpus}
Finally, we evaluated our system on {\SCICORP} corpus, 
in which, like in the {\BASHI} corpus,  gold mentions are not annotated, so again
we used NPs as our predicted mentions.
{\SCICORP}  consists of scientific documents that are very different 
from the {\BASHI} (news). 
As a result, the only reported result on  {\SCICORP}, \cite{roesiger-etal-2018-bridging}, is rather poor. 
Roesiger et al.'s rule-based system only achieved 1.5\% and 8.1\% (F1) for {\FBR} and {\AR} respectively. 
The poor result is manly due to the system only recognizing less than 1\% of the bridging anaphors,
which is another example
of the sensitivity of 
rule-based systems to domain shifting. 
By contrast, our system achieved on this corpus F1 scores of 21.8\% and 46.3\% for {\FBR} and {\AR}, respectively (Table \ref{table:all_full_bridging}). 
These scores on  {\SCICORP} are broadly in the same range to the scores achieved by our system on the other three corpora, which indicates that our system's performance doesn't deteriorate so badly with
domain shifting. 
In terms of the {\AS} task, our system achieved an accuracy of 33.4\%; to the best of our knowledge, this is the first result for {\AS} on {\SCICORP} .

\section{Conclusions}

In this paper we proposed  a multi-task neural architecture tackling two major challenges for bridging reference resolution. 
The first challenge is the lack of very large training datasets, 
as the largest corpus for bridging reference, {\ARRAU} \cite{uryupina-et-al:NLEJ}, only contains 5.5k examples, and other corpora are much smaller (the most used corpus for bridging, {\ISNOTES} \cite{markert-etal-2012-isnotes}, only contains 663 bridging pairs). 
The second challenge is that  different annotation schemes for bridging are used in different corpora,
so designing a system that can be applied to different corpora is complicated. 
Our results on the {\ARRAU} {\RST} corpus demonstrate that the performance on {\FBR} and its subtasks can be significantly improved by learning with additional coreference annotations. 
Our multi-task model achieved substantial improvements of 7.3\%-13.7\% for {\FBR} and its subtasks when compared with the single task baseline that learns solely on bridging annotations. 
As a result, our final system achieved SoTA results in all three tasks. 
Further evaluation on {\TRAINS}, {\PEAR}, {\ISNOTES}, {\BASHI} and {\SCICORP} demonstrates the robustness of our system under changes of annotation scheme and domain. 
The very same architecture used for {\ARRAU} {\RST} again achieved SoTA results on {\FBR} for all five corpora. 

Overall, our results suggest that coreference is a useful source task for bridging reference resolution, and our neural bridging architecture is  applicable to bridging corpora based on different domain or definitions of bridging.

\section*{Acknowledgements}
This research was supported in part by the DALI project, ERC Grant 695662.


\bibliographystyle{coling}
\bibliography{coling2020}

\begin{thebibliography}{}

\bibitem[\protect\citename{Baumann and Riester}2012]{baumann&riester:12}
Stefan Baumann and Arndt Riester.
\newblock 2012.
\newblock Referential and lexical givenness: semantic, prosodic and cognitive
  aspects.
\newblock In {\em Prosody and Meaning}.

\bibitem[\protect\citename{Clark and Manning}2016]{clark2016improving}
Kevin Clark and Christopher~D. Manning.
\newblock 2016.
\newblock Improving coreference resolution by learning entity-level distributed
  representations.
\newblock In {\em Proceedings of the 54th Annual Meeting of the Association for
  Computational Linguistics (Volume 1: Long Papers)}, pages 643--653, Berlin,
  Germany, August. Association for Computational Linguistics.

\bibitem[\protect\citename{Clark \bgroup et al.\egroup
  }2019]{clark-etal-2019-bam}
Kevin Clark, Minh-Thang Luong, Urvashi Khandelwal, Christopher~D. Manning, and
  Quoc~V. Le.
\newblock 2019.
\newblock {BAM}! born-again multi-task networks for natural language
  understanding.
\newblock In {\em Proceedings of the 57th Annual Meeting of the Association for
  Computational Linguistics}, pages 5931--5937, Florence, Italy, July.
  Association for Computational Linguistics.

\bibitem[\protect\citename{Clark}1975]{clark-1975-bridging}
Herbert~H. Clark.
\newblock 1975.
\newblock Bridging.
\newblock In {\em Theoretical Issues in Natural Language Processing}.

\bibitem[\protect\citename{Collobert and Weston}2008]{collobert2008unified}
Ronan Collobert and Jason Weston.
\newblock 2008.
\newblock A unified architecture for natural language processing: Deep neural
  networks with multitask learning.
\newblock In {\em ICML}.

\bibitem[\protect\citename{Cotterell and Duh}2017]{cotterell2017low}
Ryan Cotterell and Kevin Duh.
\newblock 2017.
\newblock Low-resource named entity recognition with cross-lingual,
  character-level neural conditional random fields.
\newblock In {\em Proceedings of the Eighth International Joint Conference on
  Natural Language Processing (Volume 2: Short Papers)}, pages 91--96, Taipei,
  Taiwan, November. Asian Federation of Natural Language Processing.

\bibitem[\protect\citename{Devlin \bgroup et al.\egroup }2019]{devlin2019bert}
Jacob Devlin, Ming-Wei Chang, Kenton Lee, and Kristina Toutanova.
\newblock 2019.
\newblock {BERT}: Pre-training of deep bidirectional transformers for language
  understanding.
\newblock In {\em Proceedings of the 2019 Conference of the North {A}merican
  Chapter of the Association for Computational Linguistics: Human Language
  Technologies, Volume 1 (Long and Short Papers)}, pages 4171--4186,
  Minneapolis, Minnesota, June. Association for Computational Linguistics.

\bibitem[\protect\citename{Garnham}2001]{garnham:book}
Alan Garnham.
\newblock 2001.
\newblock {\em Mental models and the interpretation of anaphora}.
\newblock Psychology Press.

\bibitem[\protect\citename{Hobbs \bgroup et al.\egroup
  }1993]{hobbs-et-al:abduction}
Jerry~R. Hobbs, Mark Stickel, Doug Appelt, and Paul Martin.
\newblock 1993.
\newblock Interpretation as abduction.
\newblock {\em Artificial Intelligence Journal}, 63:69--142.

\bibitem[\protect\citename{Hou \bgroup et al.\egroup
  }2013]{hou-etal-2013-global}
Yufang Hou, Katja Markert, and Michael Strube.
\newblock 2013.
\newblock Global inference for bridging anaphora resolution.
\newblock In {\em Proceedings of the 2013 Conference of the North {A}merican
  Chapter of the Association for Computational Linguistics: Human Language
  Technologies}, pages 907--917, Atlanta, Georgia, June. Association for
  Computational Linguistics.

\bibitem[\protect\citename{Hou \bgroup et al.\egroup
  }2014]{hou-etal-2014-emnlp}
Yufang Hou, Katja Markert, and Michael Strube.
\newblock 2014.
\newblock A rule-based system for unrestricted bridging resolution: Recognizing
  bridging anaphora and finding links to antecedents.
\newblock In {\em Proceedings of the 2014 Conference on Empirical Methods in
  Natural Language Processing ({EMNLP})}, pages 2082--2093, Doha, Qatar,
  October. Association for Computational Linguistics.

\bibitem[\protect\citename{Hou \bgroup et al.\egroup }2018]{hou-et-al:CL18}
Yufang Hou, Katja Markert, and Michael Strube.
\newblock 2018.
\newblock Unrestricted bridging resolution.
\newblock {\em Computational Linguistics}, 44(2):237--284, June.

\bibitem[\protect\citename{Hou}2016]{hou-2016-incremental}
Yufang Hou.
\newblock 2016.
\newblock Incremental fine-grained information status classification using
  attention-based {LSTM}s.
\newblock In {\em Proceedings of {COLING} 2016, the 26th International
  Conference on Computational Linguistics: Technical Papers}, pages 1880--1890,
  Osaka, Japan, December. The COLING 2016 Organizing Committee.

\bibitem[\protect\citename{Hou}2018a]{hou-2018-emnlp}
Yufang Hou.
\newblock 2018a.
\newblock A deterministic algorithm for bridging anaphora resolution.
\newblock In {\em Proceedings of the 2018 Conference on Empirical Methods in
  Natural Language Processing}, pages 1938--1948, Brussels, Belgium,
  October-November. Association for Computational Linguistics.

\bibitem[\protect\citename{Hou}2018b]{hou-2018-naacl}
Yufang Hou.
\newblock 2018b.
\newblock Enhanced word representations for bridging anaphora resolution.
\newblock In {\em Proceedings of the 2018 Conference of the North {A}merican
  Chapter of the Association for Computational Linguistics: Human Language
  Technologies, Volume 2 (Short Papers)}, pages 1--7, New Orleans, Louisiana,
  June. Association for Computational Linguistics.

\bibitem[\protect\citename{Hou}2020]{hou-2020-acl}
Yufang Hou.
\newblock 2020.
\newblock Bridging anaphora resolution as question answering.
\newblock In {\em Proceedings of the 58th Annual Meeting of the Association for
  Computational Linguistics}, pages 1428--1438, Online, July. Association for
  Computational Linguistics.

\bibitem[\protect\citename{Kamp and Reyle}1993]{kamp&reyle:93}
Hans Kamp and Uwe Reyle.
\newblock 1993.
\newblock {\em From Discourse to Logic}.
\newblock D. Reidel, Dordrecht.

\bibitem[\protect\citename{Kantor and
  Globerson}2019]{kantor-globerson-2019-coreference}
Ben Kantor and Amir Globerson.
\newblock 2019.
\newblock Coreference resolution with entity equalization.
\newblock In {\em Proceedings of the 57th Annual Meeting of the Association for
  Computational Linguistics}, pages 673--677, Florence, Italy, July.
  Association for Computational Linguistics.

\bibitem[\protect\citename{Karttunen}1976]{karttunen:76}
Lauri Karttunen.
\newblock 1976.
\newblock Discourse referents.
\newblock In {\em Syntax and Semantics 7 - Notes from the Linguistic
  Underground}, pages 363--385. Academic Press, New York.

\bibitem[\protect\citename{Kiperwasser and
  Ballesteros}2018]{kiperwasser2018scheduled}
Eliyahu Kiperwasser and Miguel Ballesteros.
\newblock 2018.
\newblock Scheduled multi-task learning: From syntax to translation.
\newblock {\em Transactions of the Association for Computational Linguistics},
  6:225--240.

\bibitem[\protect\citename{Kolhatkar \bgroup et al.\egroup
  }2018]{kolhatkar-et-al:CL18}
Varada Kolhatkar, Adam Roussel, Stefanie Dipper, and Heike Zinsmeister.
\newblock 2018.
\newblock Anaphora with non-nominal antecedents in computational linguistics: a
  {S}urvey.
\newblock {\em Computational Linguistics}, 44(3):547--612.

\bibitem[\protect\citename{Lassalle and Denis}2011]{lassalle&denis:DAARC11}
Emmanuel Lassalle and Pascal Denis.
\newblock 2011.
\newblock Leveraging different meronym discovery methods for bridging
  resolution in french.
\newblock In {\em Proc. of 8th DAARC}, pages 35--46, Faro, Portugal.

\bibitem[\protect\citename{Lee \bgroup et al.\egroup }2017]{lee2017end}
Kenton Lee, Luheng He, Mike Lewis, and Luke Zettlemoyer.
\newblock 2017.
\newblock End-to-end neural coreference resolution.
\newblock In {\em Proceedings of the 2017 Conference on Empirical Methods in
  Natural Language Processing}, pages 188--197, Copenhagen, Denmark, September.
  Association for Computational Linguistics.

\bibitem[\protect\citename{Lee \bgroup et al.\egroup }2018]{lee2018higher}
Kenton Lee, Luheng He, and Luke Zettlemoyer.
\newblock 2018.
\newblock Higher-order coreference resolution with coarse-to-fine inference.
\newblock In {\em Proceedings of the 2018 Conference of the North {A}merican
  Chapter of the Association for Computational Linguistics: Human Language
  Technologies, Volume 2 (Short Papers)}, pages 687--692, New Orleans,
  Louisiana, June. Association for Computational Linguistics.

\bibitem[\protect\citename{Luong \bgroup et al.\egroup }2016]{luong2015multi}
Minh-Thang Luong, Quoc~V Le, Ilya Sutskever, Oriol Vinyals, and Lukasz Kaiser.
\newblock 2016.
\newblock Multi-task sequence to sequence learning.
\newblock {\em ICLR}.

\bibitem[\protect\citename{Markert \bgroup et al.\egroup
  }2012]{markert-etal-2012-isnotes}
Katja Markert, Yufang Hou, and Michael Strube.
\newblock 2012.
\newblock Collective classification for fine-grained information status.
\newblock In {\em Proceedings of the 50th Annual Meeting of the Association for
  Computational Linguistics (Volume 1: Long Papers)}, pages 795--804, Jeju
  Island, Korea, July. Association for Computational Linguistics.

\bibitem[\protect\citename{Pennington \bgroup et al.\egroup
  }2014]{pennington2014glove}
Jeffrey Pennington, Richard Socher, and Christopher Manning.
\newblock 2014.
\newblock {G}lo{V}e: Global vectors for word representation.
\newblock In {\em Proceedings of the 2014 Conference on Empirical Methods in
  Natural Language Processing ({EMNLP})}, pages 1532--1543, Doha, Qatar,
  October. Association for Computational Linguistics.

\bibitem[\protect\citename{Peters \bgroup et al.\egroup }2018]{peters2018elmo}
Matthew Peters, Mark Neumann, Mohit Iyyer, Matt Gardner, Christopher Clark,
  Kenton Lee, and Luke Zettlemoyer.
\newblock 2018.
\newblock Deep contextualized word representations.
\newblock In {\em Proceedings of the 2018 Conference of the North {A}merican
  Chapter of the Association for Computational Linguistics: Human Language
  Technologies, Volume 1 (Long Papers)}, pages 2227--2237, New Orleans,
  Louisiana, June. Association for Computational Linguistics.

\bibitem[\protect\citename{Poesio and Artstein}2008]{poesio_anaphoric_2008}
Massimo Poesio and Ron Artstein.
\newblock 2008.
\newblock Anaphoric annotation in the {ARRAU} corpus.
\newblock In {\em Proceedings of the Sixth International Conference on Language
  Resources and Evaluation ({LREC}'08)}, Marrakech, Morocco, May. European
  Language Resources Association (ELRA).

\bibitem[\protect\citename{Poesio \bgroup et al.\egroup
  }2004]{poesio-etal-2004-learning}
Massimo Poesio, Rahul Mehta, Axel Maroudas, and Janet Hitzeman.
\newblock 2004.
\newblock Learning to resolve bridging references.
\newblock In {\em Proceedings of the 42nd Annual Meeting of the Association for
  Computational Linguistics ({ACL}-04)}, pages 143--150, Barcelona, Spain,
  July.

\bibitem[\protect\citename{Poesio \bgroup et al.\egroup
  }2016]{poesio-stuckardt-versley:book}
Massimo Poesio, Roland Stuckardt, and Yannick Versley.
\newblock 2016.
\newblock {\em Anaphora Resolution: Algorithms, Resources and Applications}.
\newblock Springer, Berlin.

\bibitem[\protect\citename{Poesio \bgroup et al.\egroup }2018]{poesio2018crac}
Massimo Poesio, Yulia Grishina, Varada Kolhatkar, Nafise Moosavi, Ina Roesiger,
  Adam Roussel, Fabian Simonjetz, Alexandra Uma, Olga Uryupina, Juntao Yu, and
  Heike Zinsmeister.
\newblock 2018.
\newblock Anaphora resolution with the {ARRAU} corpus.
\newblock In {\em Proceedings of the First Workshop on Computational Models of
  Reference, Anaphora and Coreference}, pages 11--22, New Orleans, Louisiana,
  June. Association for Computational Linguistics.

\bibitem[\protect\citename{Poesio}2004]{poesio:ACL_discanno04}
Massimo Poesio.
\newblock 2004.
\newblock Discourse annotation and semantic annotation in the {GNOME} corpus.
\newblock In {\em Proc. of the {ACL} Workshop on Discourse Annotation}.

\bibitem[\protect\citename{Poesio}2016]{poesio:ana_book_linguistic}
Massimo Poesio.
\newblock 2016.
\newblock Linguistic and cognitive evidence about anaphora.
\newblock In M.~Poesio, R.~Stuckardt, and Y.~Versley, editors, {\em Anaphora
  Resolution: Algorithms, Resources and Applications}, chapter~2. Springer.

\bibitem[\protect\citename{Pradhan \bgroup et al.\egroup
  }2012]{pradhan2012conllst}
Sameer Pradhan, Alessandro Moschitti, Nianwen Xue, Olga Uryupina, and Yuchen
  Zhang.
\newblock 2012.
\newblock {CoNLL-2012} shared task: Modeling multilingual unrestricted
  coreference in {OntoNotes}.
\newblock In {\em {Proceedings of the Sixteenth Conference on Computational
  Natural Language Learning (CoNLL 2012)}}, Jeju, Korea.

\bibitem[\protect\citename{Prince}1981]{prince:81}
Ellen~F. Prince.
\newblock 1981.
\newblock Toward a taxonomy of given-new information.
\newblock In P.~Cole, editor, {\em Radical Pragmatics}, pages 223--256.
  Academic Press, New York.

\bibitem[\protect\citename{Roesiger \bgroup et al.\egroup
  }2018]{roesiger-etal-2018-bridging}
Ina Roesiger, Arndt Riester, and Jonas Kuhn.
\newblock 2018.
\newblock Bridging resolution: Task definition, corpus resources and rule-based
  experiments.
\newblock In {\em Proceedings of the 27th International Conference on
  Computational Linguistics}, pages 3516--3528, Santa Fe, New Mexico, USA,
  August. Association for Computational Linguistics.

\bibitem[\protect\citename{Roesiger}2018]{roesiger-2018-crac}
Ina Roesiger.
\newblock 2018.
\newblock Rule- and learning-based methods for bridging resolution in the
  {ARRAU} corpus.
\newblock In {\em Proceedings of the First Workshop on Computational Models of
  Reference, Anaphora and Coreference}, pages 23--33, New Orleans, Louisiana,
  June. Association for Computational Linguistics.

\bibitem[\protect\citename{R{\"{o}}siger}2016]{roesiger-2016-scicorp}
Ina R{\"{o}}siger.
\newblock 2016.
\newblock {S}ci{C}orp: A corpus of {E}nglish scientific articles annotated for
  information status analysis.
\newblock In {\em Proceedings of the Tenth International Conference on Language
  Resources and Evaluation ({LREC}'16)}, pages 1743--1749, Portoro{\v{z}},
  Slovenia, May. European Language Resources Association (ELRA).

\bibitem[\protect\citename{R{\"o}siger}2018]{rosiger-2018-bashi}
Ina R{\"o}siger.
\newblock 2018.
\newblock {BASHI}: A corpus of {W}all {S}treet {J}ournal articles annotated
  with bridging links.
\newblock In {\em Proceedings of the Eleventh International Conference on
  Language Resources and Evaluation ({LREC} 2018)}, Miyazaki, Japan, May.
  European Language Resources Association (ELRA).

\bibitem[\protect\citename{Sidner}1979]{sidner:thesis}
Candace~L. Sidner.
\newblock 1979.
\newblock {\em Towards a computational theory of definite anaphora
  comprehension in English discourse}.
\newblock {Ph.D.} thesis, MIT.

\bibitem[\protect\citename{Uryupina \bgroup et al.\egroup
  }2019]{uryupina-et-al:NLEJ}
Olga Uryupina, Ron Artstein, Antonella Bristot, Federica Cavicchio, Francesca
  Delogu, Kepa~J. Rodriguez, and Massimo Poesio.
\newblock 2019.
\newblock Annotating a broad range of anaphoric phenomena, in a variety of
  genres: the {ARRAU} corpus.
\newblock {\em Journal of Natural Language Engineering}.

\bibitem[\protect\citename{Vieira and Poesio}2000]{vieira&poesio:CL}
Renata Vieira and Massimo Poesio.
\newblock 2000.
\newblock An empirically-based system for processing definite descriptions.
\newblock {\em Computational Linguistics}, 26(4):539--593, December.

\bibitem[\protect\citename{Webber}1979]{webber:thesis}
Bonnie~L. Webber.
\newblock 1979.
\newblock {\em A Formal Approach to Discourse Anaphora}.
\newblock Garland, New York.

\bibitem[\protect\citename{Webber}1991]{webber:91}
Bonnie~L. Webber.
\newblock 1991.
\newblock Structure and ostension in the interpretation of discourse deixis.
\newblock {\em Language and Cognitive Processes}, 6(2):107--135.

\bibitem[\protect\citename{Wiseman \bgroup et al.\egroup
  }2015]{wiseman2015learning}
Sam Wiseman, Alexander~M. Rush, Stuart Shieber, and Jason Weston.
\newblock 2015.
\newblock Learning anaphoricity and antecedent ranking features for coreference
  resolution.
\newblock In {\em Proceedings of the 53rd Annual Meeting of the Association for
  Computational Linguistics and the 7th International Joint Conference on
  Natural Language Processing (Volume 1: Long Papers)}, pages 1416--1426,
  Beijing, China, July. Association for Computational Linguistics.

\bibitem[\protect\citename{Wiseman \bgroup et al.\egroup
  }2016]{wiseman2016learning}
Sam Wiseman, Alexander~M. Rush, and Stuart~M. Shieber.
\newblock 2016.
\newblock Learning global features for coreference resolution.
\newblock In {\em Proceedings of the 2016 Conference of the North {A}merican
  Chapter of the Association for Computational Linguistics: Human Language
  Technologies}, pages 994--1004, San Diego, California, June. Association for
  Computational Linguistics.

\bibitem[\protect\citename{Yang \bgroup et al.\egroup }2017]{yang2017transfer}
Zhilin Yang, Ruslan Salakhutdinov, and William~W Cohen.
\newblock 2017.
\newblock Transfer learning for sequence tagging with hierarchical recurrent
  networks.
\newblock {\em ICLR}.

\bibitem[\protect\citename{Zhou \bgroup et al.\egroup }2019]{zhou2019dual}
Joey~Tianyi Zhou, Hao Zhang, Di~Jin, Hongyuan Zhu, Meng Fang, Rick Siow~Mong
  Goh, and Kenneth Kwok.
\newblock 2019.
\newblock Dual adversarial neural transfer for low-resource named entity
  recognition.
\newblock In {\em Proceedings of the 57th Annual Meeting of the Association for
  Computational Linguistics}, pages 3461--3471, Florence, Italy, July.
  Association for Computational Linguistics.

\end{thebibliography}

\end{document}